\documentclass{article}

\usepackage{arxiv}

\usepackage[utf8]{inputenc} % allow utf-8 input
\usepackage[T1]{fontenc}    % use 8-bit T1 fonts
\usepackage{hyperref}       % hyperlinks
\usepackage{url}            % simple URL typesetting
\usepackage{booktabs}       % professional-quality tables
\usepackage{amsfonts}       % blackboard math symbols
\usepackage{nicefrac}       % compact symbols for 1/2, etc.
\usepackage{microtype}      % microtypography
\usepackage{lipsum}		% Can be removed after putting your text content
\usepackage{graphicx}
\usepackage{natbib}
\usepackage{doi}
\usepackage{siunitx}

\title{Vision Xformers: Efficient Attention for Image Classification }

\author{ \href{https://orcid.org/0000-0003-4110-9638}{\includegraphics[scale=0.06]{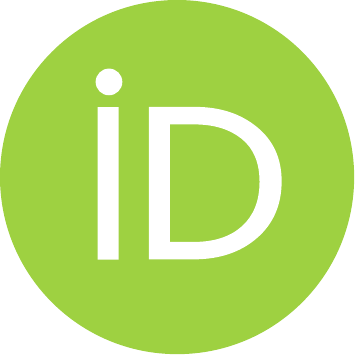}\hspace{1mm}Pranav Jeevan P} \\
	Department of Electrical Engineering\\
	Indian Institute of Technology Bombay\\
	Mumbai, India  \\
	\texttt{194070025@iitb.ac.in} \\
	\And
	\href{https://orcid.org/0000-0002-8003-6809}{\includegraphics[scale=0.06]{orcid.pdf}\hspace{1mm}Amit Sethi} \\
	Department of Electrical Engineering\\
	Indian Institute of Technology Bombay\\
	Mumbai, India \\
	\texttt{asethi@iitb.ac.in} \\
}

\date{}

\renewcommand{\headeright}{Preprint}
\renewcommand{\undertitle}{Preprint}
\renewcommand{\shorttitle}{Vision Xformers}

\hypersetup{
pdftitle={Vision Xformers: Efficient Attention for Image Classification},
pdfsubject={q-bio.NC, q-bio.QM},
pdfauthor={Pranav Jeevan P, Amit Sethi},
pdfkeywords={Transformer, Image classification, Linear Attention, Deep Learning},
}

\begin{document}
\maketitle

\begin{abstract}
Although transformers have become the neural architectures of choice for natural language processing, they require orders of magnitude more training data, GPU memory, and computations in order to compete with convolutional neural networks for computer vision. The attention mechanism of transformers scales quadratically with the length of the input sequence, and unrolled images have long sequence lengths. Plus, transformers lack an inductive bias that is appropriate for images. We tested three modifications to vision transformer (ViT) architectures that address these shortcomings. Firstly, we alleviate the quadratic bottleneck by using linear attention mechanisms, called X-formers (such that, X $\in$ \{Performer, Linformer, Nyströmformer\}), thereby creating Vision X-formers (ViXs). This resulted in up to a seven times reduction in the GPU memory requirement. We also compared their performance with FNet and multi-layer perceptron mixers, which further reduced the GPU memory requirement. Secondly, we introduced an inductive bias for images by replacing the initial linear embedding layer by convolutional layers in ViX, which significantly increased classification accuracy without increasing the model size. Thirdly, we replaced the learnable 1D position embeddings in ViT with Rotary Position Embedding (RoPE), which increases the classification accuracy for the same model size. We believe that incorporating such changes can democratize transformers by making them accessible to those with limited data and computing resources.

\end{abstract}

% keywords can be removed
\keywords{Transformer \and Image classification \and Linear Attention \and Deep Learning}

\section{Introduction}

Transformers have revolutionised the natural language processing (NLP) domain with their ability to handle context over long sequences of data and achieving human-level accuracy for various tasks, such as language translation, text summarization, question answering, language modeling, and text generation \cite{vaswani2017attention,devlin2019bert}. On the other hand, in recent years, vision application have been almost completely dominated by the convolutional neural network (CNN) architectures, which can exploit the two-dimensional (2D) structure of images using inductive priors, such as translational equivariance due to convolutional weight sharing and partial scale invariance due to pooling operations. Even though CNNs possess these advantages in handling image data, they cannot scale up receptive fields without increasing network depth and require several layers, especially pooling-type mechanisms, to capture long range dependencies. While the effective weights are dynamically calculated based on the inputs in attention mechanisms for transformers, popular CNN architectures, such as ResNet \cite{he2015deep} and InceptionNet \cite{szegedy2014going} lack an attention mechanism and use static weights for each input, although there are exceptions, such as squeeze and excitation nets \cite{iandola2016squeezenet,hu2019squeezeandexcitation}. Thus, CNNs can further benefit from carefully crafted attention mechanisms to capture information from all spatial locations of an input. These advantages of attention mechanisms over vanilla CNNs have motivated the research into transformer architectures suitable for vision applications~\cite{khan2021transformers}.

In spite of their advantages over CNNs, transformers with their default attention mechanisms have found limited use in vision due to their quadratic complexity with respect to sequence length. Images are fed into transformers as long unrolled 1D sequences of pixels or patches. For example, an image of size 256×256 pixels becomes a sequence of length 65,536 pixels. The computational and memory requirements also scale quadratically as we use images of higher resolutions. Transformers also lack inductive biases to exploit the 2D structure of the images. A general architecture without a strong inductive bias makes transformers suitable for applications in multiple domains, provided one has enough data and computational resources. However, the required data, resources, and time for training transformers to achieve accuracy comparable performance with CNNs on images is not suitable for resource-constrained scenarios \cite{khan2021transformers}. The inductive bias of 2D convolutions in the latter allows them to learn from relatively smaller datasets.

We propose three specific improvements to vision transformers to reduce the resources required to process images: (i) replacing the quadratic attention in ViT with linear attention mechanisms such as Linformer \cite{wang2020linformer}, Performer \cite{choromanski2021rethinking} and Nyströmformer \cite{xiong2021nystromformer} creating the Vision X-former (ViX), (ii) replacing the the initial linear embedding layer by convolutional layers in ViX creating a hybrid ViX, and (iii) replacing the popular learnable 1D position embedding with Rotary Position Embedding (RoPE) \cite{su2021roformer}. While, individually, some of the sub-quadratic attention mechanisms have been tested on images, none of the testing started with images in mind for a thorough comparison of various attention mechanisms for vision. Our idea of using convolutional embedding layers was also developed in parallel by another group \cite{xiao2021early}, while the use of RoPE for vision transformers has not been proposed before. These improvements help us use smaller network sizes for images, which result in the processing of longer sequences at lesser cost. Our experiments suggest that each of these changes independently contribute to the increase in image classification accuracy on relatively smaller datasets using supervised learning. We also demonstrate these improvements across alternative transformer models, such as LeViT, CCT (Compact Convolutional Transformer), CvT  (Convolutional vision Transformer) and PiT (Pooling-based Vision Transformer) when we tested them after making the proposed changes to their architectures. We also observed similar trends in the performance of FNet and MLP mixer-based vision transformers, when these are modified as proposed.

\section{Related Works}

\subsection{Vision transformers}

\begin{figure}[h]
\centering
\includegraphics[scale=0.85]{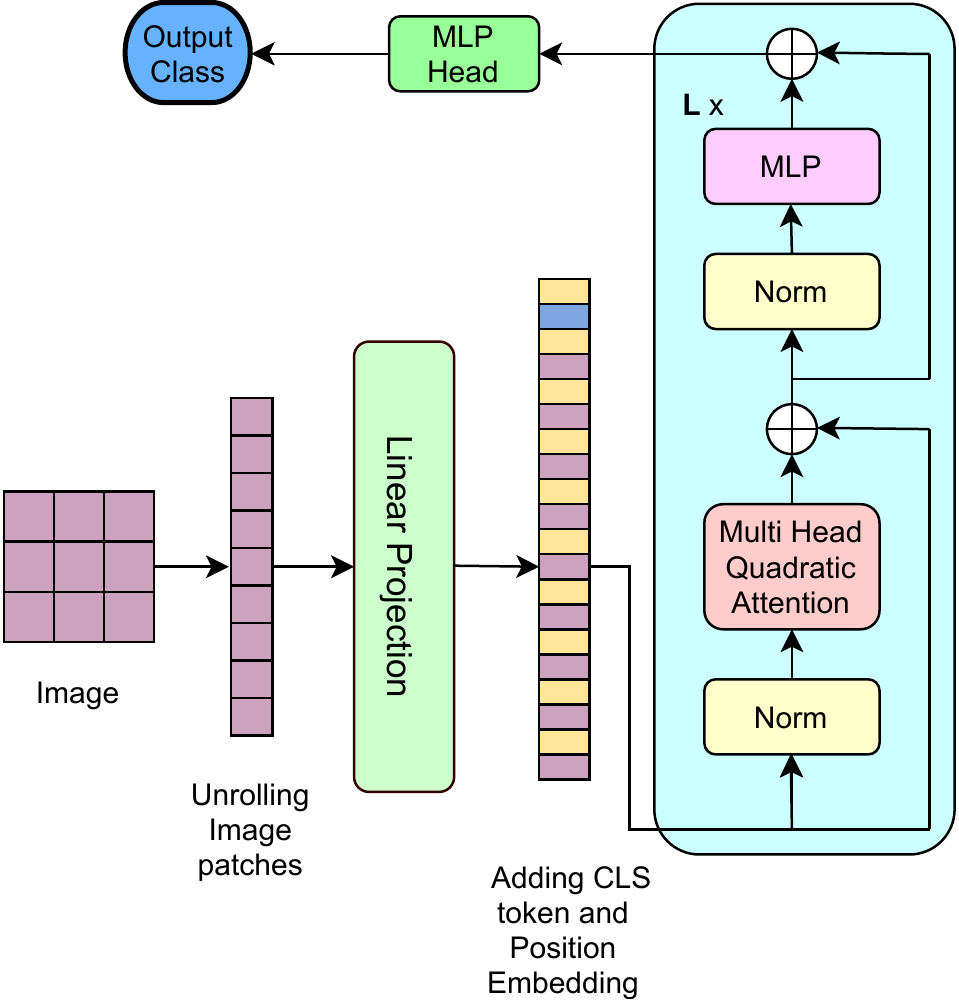}
\caption{ViT architecture}
\label{fig:x ViT}
\end{figure}

Transformers have been applied to unrolled pixel sequences~\cite{parmar2018image}. With Vision Transformer (ViT), it was proposed that applying the original transformer model to a sequence of sub-images (a.k.a. patches) rather than individual pixels  of an image shortens the sequence length \cite{dosovitskiy2021image}, as shown in Figure \ref{fig:x ViT}. Two-dimensional image patches are flattened into a vector and fed to the transformer as a sequence. These vectorized patches are then projected onto a patch embedding using a linear layer and 1D learnable position embedding is attached with it to encode location information. A learnable class embedding is prepended to the sequence of embedded patches. The output representation corresponding to this position is used as the global image representation for the image classification task. ViT achieved excellent results on image classification compared to state-of-the-art CNN models.  

However, architecture that use unrolled pixel or even patch sequences have to be pre-trained on a large dataset (e.g., 300 million images~\cite{dosovitskiy2021image}) and yet they show poor performance on medium-size datasets due to the lack of inductive biases for images. A larger number of images is required to discover the knowledge rules when a suitable inductive bias, e.g. 2D translational equivariance, is not used.

%-------------------------------------------------------------------------

%-------------------------------------------------------------------------

\subsection{Linear transformers}

Recently, there has been an influx of proposals for efficient transformers (X-formers) to tackle the problem of quadratic complexity of the number of attention weights with respect to the sequence length \cite{tay2020efficient}. These X-formers, such as Linformer \cite{wang2020linformer}, Performer  \cite{choromanski2021rethinking}, Linear Transformer \cite{katharopoulos2020transformers} and Nyströmformer \cite{xiong2021nystromformer} claim linear complexity in computing attention, which makes them ideal candidates for analyzing long pixel sequences encountered in vision tasks. The Linformer projects the length dimension of keys and values to a lower dimensional representation \cite{wang2020linformer}. Performer \cite{choromanski2021rethinking} and Linear transformer \cite{katharopoulos2020transformers} use kernels to cleverly re-write the attention mechanism to avoid computing the N×N attention matrix. Nyströmformer repurposes the Nyström method for approximating self-attention using landmark (or Nyström) points to reconstruct the softmax matrix in self-attention \cite{xiong2021nystromformer}. The Long Range Arena (LRA) benchmark was established to compare the performance and efficiency of these X-formers on various long sequence tasks, including image classification \cite{tay2020long}.
Most of the applications of linear transformers discussed above have been limited to the NLP domain. We exploit the linear complexity of these attention mechanisms to create a more efficient transformer for vision applications that can handle long sequences and smaller patch sizes with reduced GPU memory. 

\subsection{Convolutional embedding layers for ViT}

It was observed in \cite{child2019generating} and \cite{dosovitskiy2021image} that early layers in the transformer learn locally connected patterns which resemble convolutions. This led to the development of many hybrid architectures inspired by transformers and CNNs. LeViT is such a hybrid architecture for fast inference image classification using both convolutional layers and attention \cite{graham2021levit}. LeViT uses convolutional embedding instead of patch-wise projection and uses 2D relative positional biases instead of initial absolute positional bias used in ViT. It also down-samples the image in stages, uses an extra non-linearity in attention and also replaces layer-norm with batch-norm. CvT improves upon ViT by using a hierarchy of transformers containing a new convolutional token embedding, and a convolutional transformer block \cite{wu2021cvt}. CCT eliminates the requirement for class token and positional embedding through a novel sequence pooling strategy and the use of convolutions \cite{hassani2021escaping}. PiT uses the spatial dimension reduction principle of CNN to improve model capability and generalization performance compared to ViT \cite{heo2021rethinking}. LeViT, CvT and CCT uses convolution layers to down-sample the image size to reduce the sequence length. Additionally, they added multiple convolutional layers between the attention blocks, which increases the number of model parameters since CNNs have more parameters than transformers. Therefore, we used a single convolutional block in our architecture replacing the initial linear embedding layer of ViT to provide the required inductive bias to our model, without down-sampling the image size. 

%-------------------------------------------------------------------------

\subsection{Alternative positional embedding}

Rotary position embedding (RoPE) encodes the absolute positional information with a rotation matrix and incorporates an explicit relative position dependency in the self-attention mechanism \cite{su2021roformer}. It captures decaying inter-token dependency with increasing relative distances and can be used easily with linear self-attention mechanisms with relative position encoding. The previous works using RoPE has been limited to NLP and molecular domains \cite{ross2021large}. We observed the change in classification accuracy when the standard learnable 1D positional embedding is replaced with RoPE in the ViT.

\subsection{Alternative mixing models}
Recently, new architectures were introduced that replace attention with other mechanisms that enable mixing of information between tokens. FNet replaces the self-attention layer in the transformer with the Fourier Transform that does not need to be learned \cite{leethorp2021fnet}. FNet was tested mostly in NLP tasks and its performance was compared to language models like BERT. MLP-mixer replaces attention using two types of multi-layer perceptrons (MLPs) which are applied independently first to image patches and then across patches \cite{tolstikhin2021mlpmixer}. In this paper, we compare the performance and GPU consumption of these models in supervised image classification by replacing attention in ViT with these alternative mixing mechanisms. We also observe the improvement in accuracy when we add convolutional layers to create the embedding.

\section{Proposed Architectures for ViX}

\begin{figure}[h]
\centering
\includegraphics[scale=0.85]{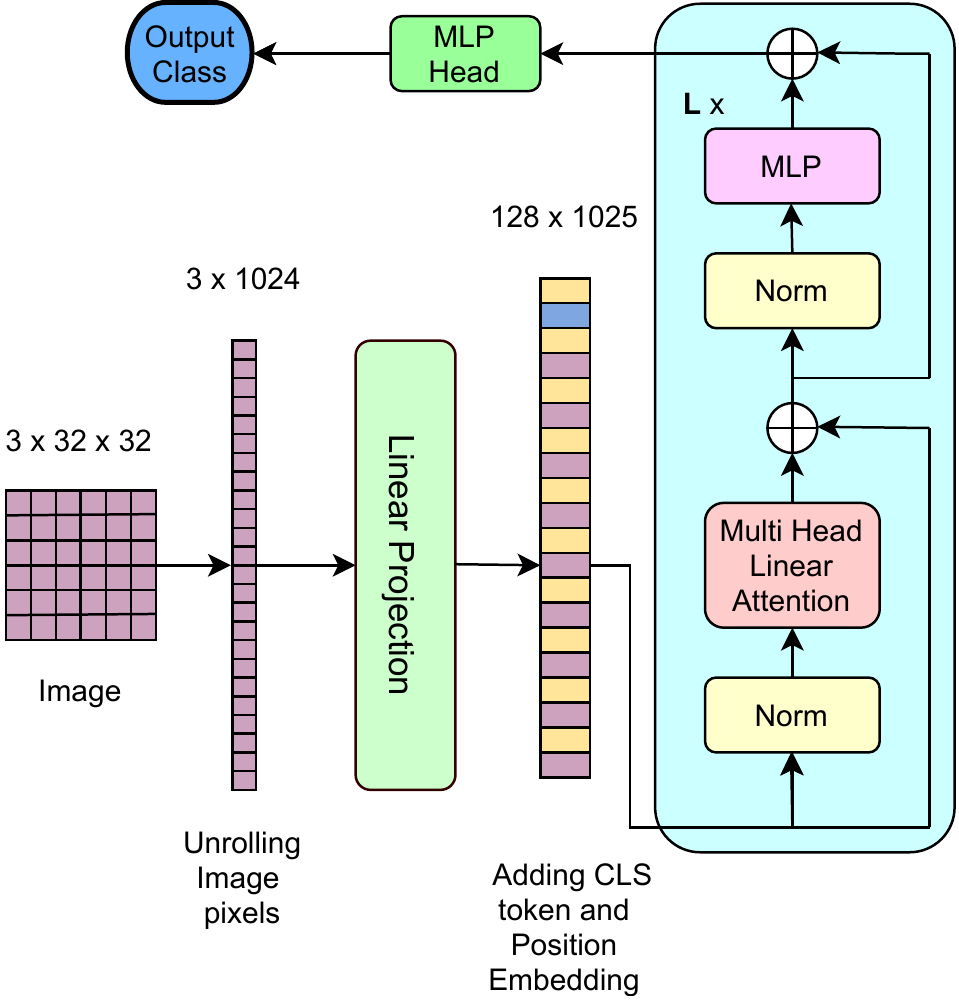}
\caption{ViX architecture}
\label{fig:x ViX}
\end{figure}

We replace the regular quadratic attention in ViT with linear attention such as Linformer, Performer and Nyströmformer, creating the Vision X-former (ViX), where X $\in$ \{Performer, Linformer, Nyströmformer\}, and compare their accuracy and computational requirements for image classification. We set the patch size of the ViT to one so that each pixel is considered as a separate token and the entire image is unrolled to create the full-length image token sequence. This is done to observe the performance in long sequence image classification. We also test the performance of FNet and MLP mixer architectures of similar depth to compare them with ViX. The ViX architecture is shown in Figure~\ref{fig:x ViX}.

\begin{figure}[h]
\centering
\includegraphics[scale=0.64]{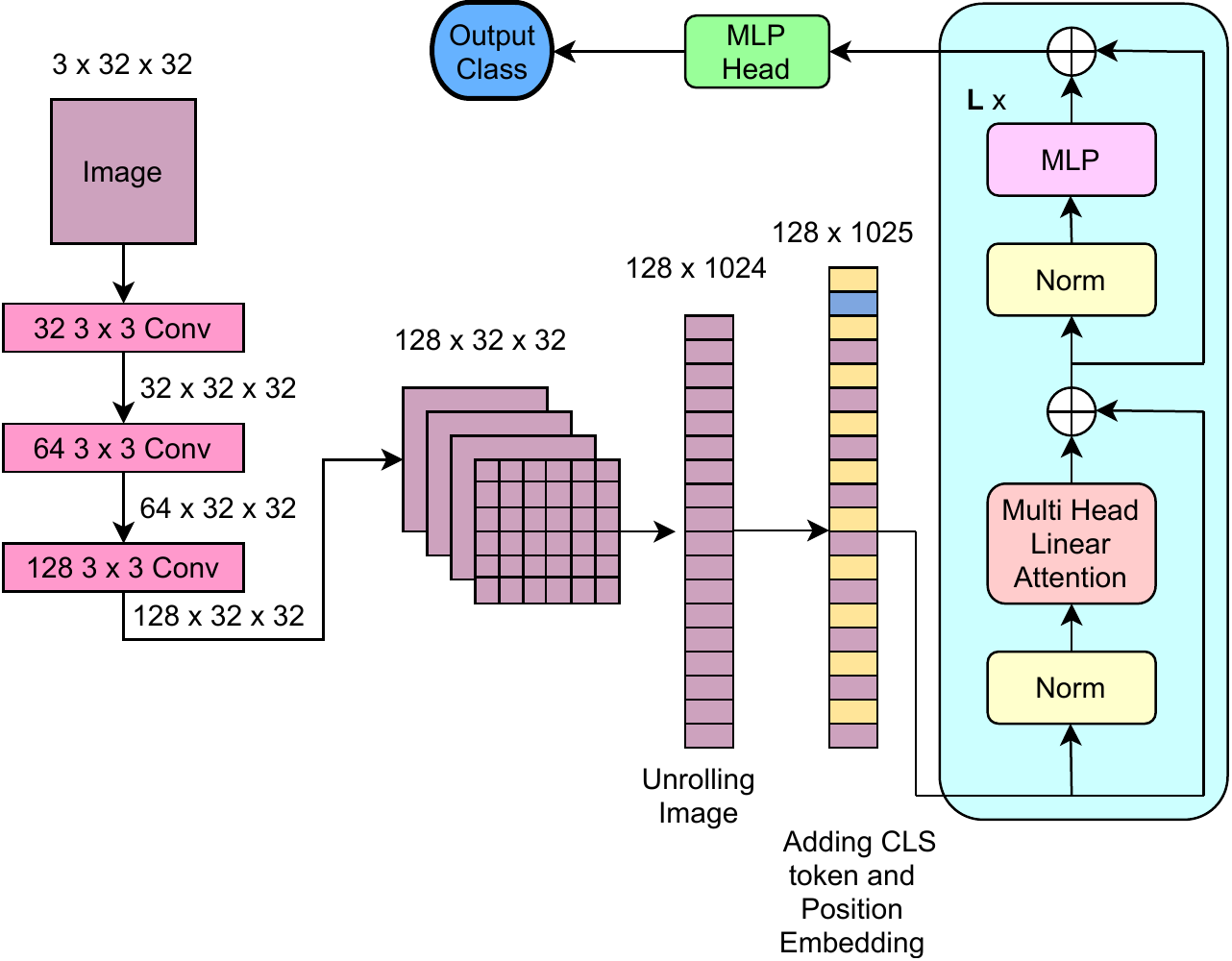}
\caption{Hybrid ViX architecture }
\label{fig:x HybridViX}
\end{figure}

The lack of inductive priors restricts attention from exploiting the 2D nature and symmetries of the images in the low data regime. We therefore use convolutional layers to replace the initial linear embedding layer to extract 2D information from the images. We propose hybrid ViX -- a combination of linear attention and convolutional layers for image classification -- and compare its performance with standard ViT. We replace the linear embedding layer in ViX by three convolutional layers, creating the same number of feature maps as the embedding dimension. The size of each feature map is the same as that of the input image. The hybrid ViX architecture is shown in Figure \ref{fig:x HybridViX}. The convolutional layers were also added to FNet and MLP mixer models to observe their performance.

We also compare the performance of other transformer architectures such as LeViT, CvT, PiT, and CCT by replacing quadratic attention with linear ones, creating LeViX, CCX, CvX, and PiX. In our notation, we replace the X in X-former models with the name of the linear attention used, for example, CCT model that uses Performer becomes CCP (Compact Convolutional Performer). Addition of the prefix \emph{hybrid} means that the convolutional layer has replaced the linear embedding layer, i.e, hybrid PiX is created by replacing the linear embedding layer with convolutional ones. LeViT adds relative positional bias when it computes the full attention matrix. Since, linear attention mechanism does not compute this matrix, we could not add the relative positional bias in LeViX.  Therefore, we removed the positional bias in LeViT so that the comparison between LeViT and LeViX is justified. Subsequently, we compare the performance of ViX and hybrid ViX models by replacing 1D learnable position embedding with a rotary positional embedding.

\section{Experiments}

The CIFAR-10~\cite{Krizhevsky09learningmultiple} and the Tiny ImageNet~\cite{Le2015TinyIV} datasets were used for our experiments. Since the image size in the former is 32×32, we have a sequence length of 1024 upon unrolling the images and an additional class token was added for classification, making the final sequence length 1025. All models used in our experiments have 128-dimensional embedding, 4 layers, 4 heads, 256 as MLP dimension, and 0 dropout to ensure uniformity in comparison in line with previous studies.\footnote{The codes for the implementation of all models were adapted from \href{https://github.com/lucidrains/vit-pytorch}{https://github.com/lucidrains/vit-pytorch}} We used 64 benchmark points in Nyströmformer, a local window size of 256 with ReLU nonlinearity as the kernel function in Performer, and used the same projection for keys and values in Linformer. The Adam optimizer with $\alpha = 0.001$ (learning rate), $\beta_{1} = 0.9$ and $\beta_{2}=0.999$ were used for computing running averages of gradient and its square, $\epsilon = 10^{-8}$, and 0.01 as weight decay coefficient were used. We used automatic mixed precision in PyTorch during training to make it faster and consume less memory. Experiments were done with 16 GB Tesla P100-PCIe and Tesla T4 GPUs available in Kaggle and Google Colab. GPU usage for a batch size of 64 is reported along with top-1 \% and top-5\% accuracy from best of three runs.

In the hybrid ViT and ViX models, we used three convolutional layers with 32, 64, and 128 3×3 kernels respectively in each stage with stride of one with padding to generate 128 feature maps of size 32×32. We used four-head, four-layer ViT architecture with patch dimension of 128, and MLP dimension of 256 for this experiment. For the comparison between LeViT and LeViX, we used a single stage model with three convolutional layers for generating the embedding. For CCT and CCX, we used learnable position embedding and 3×3 convolutional and pooling kernels having stride equal to 1 and padding. For CvT and CvX, we used one attention module in the first stage, one attention module in the second stage and 2 attention modules in the third stage with 3×3 convolutions in each stage. For the PiT, PiX, and hybrid PiX, we used a 4-layer transformer, down-sampling the sequence after every two layers.

We also used the Tiny ImageNet dataset to validate the results obtained from CIFAR-10. We set the patch size of ViX models and stride for the initial convolutional layer in hybrid ViX models as two so that the 64×64 images in the Tiny ImageNet dataset gets unrolled to a sequence of length 1024. The final classification layer was replaced to handle classification of 200 classes instead of 10 classes. All other design parameters were similar to those used for the CIFAR-10 dataset.
%------------------------------------------------------------------------

\section{Results}

\subsection{ViX compared to ViT}

\begin{figure}[h]
\centering
\includegraphics[scale=.65]{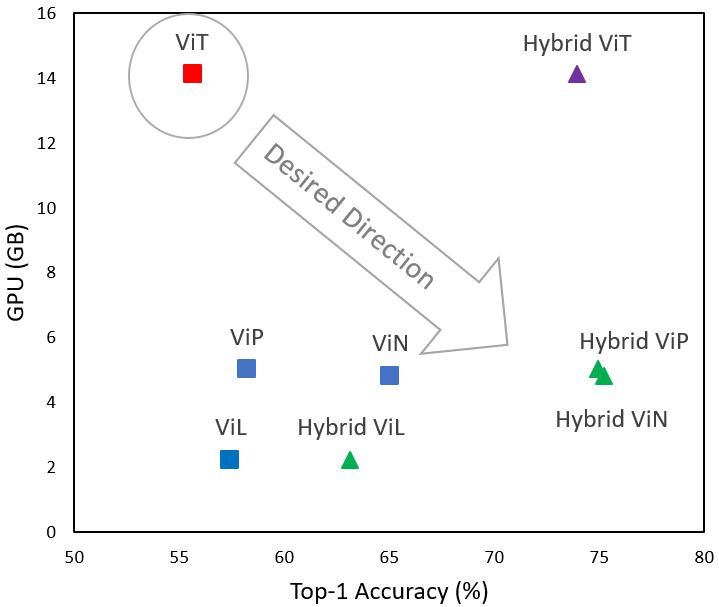}
\caption{Comparison of classification accuracy and GPU usage by various models on CIFAR-10 for a batch size of 64}
\label{fig:resultsScatter}
\end{figure}

Compared to the vanilla transformer (ViT), the Vision Nyströmformer (ViN) performs $\approx 16  \%$ better, Vision Performer (ViP) performs $\approx 3 \%$ better, and Vision Linformer (ViL) performs $\approx 2 \%$ better for image classification of CIFAR-10 dataset, as show in Table~\ref{tab:ViX}. Among the ViX architectures, ViP and ViN use almost the same number of parameters as the quadratic attention, but ViL uses a significantly lower number of parameters. ViT also requires the largest share of GPU memory and storage space among all models tested. All X-former models (blue) consume less than half the storage space used by ViT (red) and use less than one-third of ViT's GPU memory for achieving comparable performance as shown in Figure~\ref{fig:resultsScatter}. Similar improvements in performance at reduced GPU RAM consumption were also observed on classification of Tiny ImageNet dataset, as shown in Table~\ref{tab:tiny}.

%\begin{figure}[ht]
%\begin{center}
%\includegraphics[scale=0.64]{GPU.JPG}
%\caption{Comparison of GPU usage by various ViX models and ViT for batch size of 64}
%\label{fig:x GPU}
%\end{center}
%\end{figure}

% \usepackage{booktabs}
% \begin{table}

\begin{table*}[h]
\centering
\resizebox{\textwidth}{!}{%
\begin{tabular}{lrrrrr}
\hline
\textbf{Model} & \textbf{Parameters} & \textbf{Size (MB)} & \textbf{Top-1 Acc. (\%)} & \textbf{Top-5 Acc. (\%)} & \textbf{GPU (GB)} \\
\hline
Vision Transformer (ViT)& 530,442 & 209.39 & 56.81 & 94.82 & 14.1 \\ \hline
Vision Performer (ViP)& 531,978 & 85.15 & 58.23 & 94.89 & 5.0 \\
Vision Linformer (ViL)& \textbf{415,754} & \textbf{77.70} & 57.45 & 95.11 & \textbf{2.2} \\
Vision Nyströmformer (ViN)& 530,970 & 86.62 & \textbf{65.06} & \textbf{96.42} & 4.8 \\ 
\hline
FNet &  267,786 & 60.12 & 50.87 & 93.22 & 1.5 \\
MLP Mixer & 8,533,002 & 90.59 & 60.33 & 95.79 & 1.4\\\hline
Hybrid Vision Transformer (ViT)& 623,178 & 211.47 & 73.94 & 98.08 & 14.1 \\
Hybrid Vision Performer (ViP)& 624,714 & 87.23 & 74.98 & 98.41 & 5.0 \\
Hybrid Vision Linformer (ViL)& \textbf{508,490} & \textbf{79.78} & 63.17 & 95.75 & \textbf{2.2} \\
Hybrid Vision Nyströmformer (ViN)& 623,706 & 88.70 & \textbf{75.26} & \textbf{98.39} & 4.8 \\ \hline
Hybrid FNet & 360,522 & 62.20 & 64.34 & 96.95 & 1.6 \\
Hybrid MLP Mixer & 8,625,738 & 92.67 & 61.58 & 96.26 & 1.4\\\hline \\
\end{tabular}% 
}
\caption{Test accuracy and various requirements of models compared on the CIFAR-10 dataset for a batch size of 64}
\label{tab:ViX}
%\end{centering}
\end{table*}

FNet has the lowest number of parameters since it completely replaces the attention module with the fast Fourier transform that does require trainable parameters. The four-layered FNet performs $\approx 10 \%$ worse than ViT but uses only half the number of parameters, one-third of its memory, and only one-tenth of the GPU RAM for image classification on the CIFAR-10 dataset, as shown in Table~\ref{tab:ViX}. The 4 layered MLP mixer performs $\approx 7 \%$ better, but uses 16 times the number of parameters that a ViT uses. However, an MLP mixer uses less than half the memory and one-tenth of the GPU RAM used by a ViT, making it a more efficient alternative to transformers. FNet performs $\approx 4$ percentage points better than MLP mixer in classification of Tiny ImageNet dataset using one-tenth the GPU RAM used by ViT as shown in Table~\ref{tab:tiny}. The fractional GPU RAM usage by these models will enable us to use deeper models compared to ViT for a fixed GPU budget.

%-------------------------------------------------------------------------
\subsection{Hybrid ViX compared to ViX}

\begin{table*}[h]
%\begin{center}
\centering
\resizebox{\textwidth}{!}{%
%\begin{tabular}{lccccc}

\begin{tabular}{lrrrrr}
\hline
\textbf{Model} & \textbf{Parameters} & \textbf{Size (MB)} & \textbf{Top-1 Acc. (\%)} & \textbf{Top-5 Acc. (\%)} & \textbf{GPU (GB)}  \\%\multicolumn{1}{c}{\textbf{Models}} & \textbf{Parameters} & \textbf{Size (MB)} & \textbf{Top 1 Accuracy (\%)} & \textbf{Top 5 Accuracy (\%)} & \textbf{GPU (GB)} \\ 
\hline
Vision Transformer (ViT) & 556,104 & 209.59 & 26.43 & 51.82 & 14.1 \\ \hline
Vision Performer (ViP) & 557,640 & 85.35 & 28.17 & 53.93 & 4.4 \\
Vision Linformer (ViL)& \textbf{441,416} & \textbf{77.90} & 24.38 & 49.31 & \textbf{2.0} \\
Vision Nyströmformer (ViN) & 556,632 & 86.82 & \textbf{30.17} & \textbf{55.97} & 3.6 \\ \hline
FNet & 293,448 & 60.32 & 22.75 & 49.85 & 1.5 \\
MLP Mixer & 8,558,664 & 90.79 & 18.47 & 42.16 & 1.2 \\ \hline
Hybrid Vision Transformer (ViT)& 647,688 & 211.60 & \textbf{34.48} & 60.79 & 14.1 \\
Hybrid Vision Performer (ViP)& 649,224 & 87.36 & 34.13 & 60.42 & 4.4 \\
Hybrid Vision Linformer (ViL) & \textbf{533,000} & \textbf{79.91} & 26.90 & 52.88 & \textbf{2.2} \\
Hybrid Vision Nyströmformer (ViN)& 648,216 & 88.83 & 34.05 & \textbf{60.80} & 3.6 \\ \hline
Hybrid FNet & 385,032 & 62.33 & 26.39 & 51.86 & 1.6 \\
Hybrid MLP Mixer & 8,650,248 & 92.80 & 22.76 & 47.33 & 1.2 \\
\hline \\

\end{tabular}%
}
\caption{Test accuracy and various requirements of models compared on the Tiny ImageNet dataset for a batch size of 64}
\label{tab:tiny}
%\end{center}
\end{table*}
As we can see from Table~\ref{tab:ViX}, replacing the initial linear embedding layer with convolutional layers increases the number of parameters slightly for all base architectures. We also observe a slight increase in the model size and a negligible increase in the GPU usage. However, these small increases in parameters and computational resources is compensated by an appreciable improvement in performance of $\approx 20 \%$ across all models for classification of CIFAR-10 dataset. Performer (Hybrid ViP) and transformer (Hybrid ViT) models both gained $\approx 28 \%$ accuracy, while the Nyströmformer (Hybrid ViN) gained $\approx 16 \%$. The Hybrid ViP and the Hybrid ViN outperformed the Hybrid ViT by about $1$ and $2$ percentage points, respectively, using only one-third of the GPU RAM and half the memory of a ViT. Even though the Hybrid Vision Linformer (ViL) underperformed compared to the other mechanisms, its computational cost is one-sixth of ViT and it also has the lowest number of parameters. The Hybrid ViN performed better than the other models in terms of the top-1 accuracy. 

Adding convolutional layers to FNet improves its performance by $\approx 28 \%$ for CIFAR-10 dataset while it improves by $\approx 18 \%$ for Tiny ImageNet dataset. It has the smallest number of parameters and consumes less memory than other ViX models, making it an attractive alternative compared to transformer models. The classification accuracy of the MLP mixer increased only marginally with the addition of convolutional layers for CIFAR-10 dataset, but increased by 4 percentage points for Tiny ImageNet dataset. 

%\begin{figure}[ht]
%\begin{center}
%\includegraphics[scale=0.78]{graph.JPG}
%\caption{Test accuracy of various models models on CIFAR-10 dataset}
%%\label{fig:x graph}
%\end{center}
%\end{figure}

Figure \ref{fig:resultsScatter} shows clearly that all ViX (blue squares) and hybrid (triangles) models outperform the ViT (red). The linear attention of ViX models (blue) using Performer (ViP) and Nyströmformer (ViN) outperform the quadratic attention of ViT. Further, the graph shows that all hybrid models (green and purple), which use convolutional layers for generating the embedding, outperform the models that use a linear layer (red and blue) by a large margin. Even when the convolutional layers replace the linear embedding layers, the ViX models (green) outperform normal attention in ViT (purple). This significant increase in accuracy may be attributed to the convolutional layers providing low-level features such as edges, corners, and bars to the attention layers in the low data regime. The availability of these inductive priors from convolutional layers helps the transformer learn to classify images using less data. 

% \usepackage{graphicx}

Adding convolutional layers significantly improved the performance of models in classification of Tiny ImageNet dataset. We can see from Table~\ref{tab:tiny} that hybrid ViP and hybrid ViN achieve the accuracy of hybrid ViT with less than one-third the GPU RAM. These results holds potential for using hybrid X-formers for processing high resolution images at smaller patch sizes without exhausting GPU memory. 
%-------------------------------------------------------------------------

\subsection{ViX compared to other Vision Transformers}

\begin{table*}[h]
\centering
%\resizebox{\textwidth}{!}{%
\begin{tabular}{lrrrrr}
\hline
\textbf{Model} & \textbf{Parameters} & \textbf{Size (MB)} & \textbf{Top-1 Accuracy (\%)} & \textbf{Top-5 Accuracy (\%)} & \textbf{GPU (GB)} \\

\hline
LeViT & 624,970 & 213.15 & 77.06 & 98.46 & 13.9 \\
LeViP & 662,410 & 77.14 & \textbf{79.50} & \textbf{98.91} & 4.9 \\
LeViL & \textbf{506,186} & \textbf{69.69} & 63.17 & 96.13 & \textbf{2.3} \\
LeViN & 621,402 & 77.13 & 77.81 & 98.61 & 4.3 \\ \hline
CCT & 905,547 & 225.47 & 82.23 & 99.04 & 14.7 \\
CCP & 907,083 & 101.48 & 82.48 & 99.06 & 4.3 \\
CCL & \textbf{790,859} & \textbf{94.04} & 80.05 & 98.92 & \textbf{3.5} \\
CCN & 906,075 & 101.48 & \textbf{83.36} & \textbf{99.07} & 4.2 \\ \hline
CvT & 1,099,786 & 173.21 & 79.93 & 99.02 & 12.8 \\
CvP & 827,914 & 100.17 & 83.19 & \textbf{99.20} & 4.8 \\
CvL & \textbf{711,690} & \textbf{92.73} & 72.58 & 97.81 & \textbf{2.6} \\
CvN & 826,906 & 100.17 & \textbf{83.26} & 99.14 & 4.3 \\ \hline \\
\end{tabular}%
%}

\caption{Comparison of different convolutional transformers without and with linear attention mechanisms on CIFAR-10 dataset}
\label{tab:cifLeV}
\end{table*}

% \usepackage{graphicx}
\begin{table*}[h]
\centering
%\begin{center}
%\resizebox{\textwidth}{!}{%
\begin{tabular}{lrrrrr}
\hline
\textbf{Model} & \textbf{Parameters} & \textbf{Size (MB)} & \textbf{Top-1 Accuracy (\%)} & \textbf{Top-5 Accuracy (\%)} & \textbf{GPU (GB)} \\ \hline
LeViT & 649,480 & 213.28 & 38.29 & 65.23 & 14 \\
LeViP & 646,920 & 77.27 & \textbf{41.49} & \textbf{68.39} & 4.9 \\
LeViL & \textbf{530,696} & \textbf{69.89} & 26.52 & 51.85 & \textbf{3.3} \\
LeViN & 645,912 & 77.26 & 38.28 & 64.94 & 4.3 \\
\hline
CCT & 930,057 & 225.61 & 39.05 & 65.74 & 14.2 \\
CCP & 931,593 & 101.61 & 40.11 & 66.93 & 4.1 \\
CCL & \textbf{815,369} & \textbf{94.17} & 37.24 & 62.97 & \textbf{2.3} \\
CCN & 930,585 & 101.61 & \textbf{40.80} & \textbf{67.81} & 3.9 \\
\hline
CvT & 1,124,296 & 173.34 & 40.69 & 68.19 & 12.3 \\
CvP & 852,424 & 100.3 & 45.26 & 71.21 & 4.4 \\
CvL & \textbf{736,200} & \textbf{92.86} & 33.95 & 61.29 & \textbf{2.3} \\
CvN & 851,416 & 100.3 & \textbf{45.47} & \textbf{71.35} & 4.0 \\
\hline \\
\end{tabular}%
%}
%\end{center}
\caption{Comparison of different convolutional transformers without and with linear attention mechanisms on Tiny ImageNet dataset}
\label{tab:tinyLeV}
\end{table*}

% The tabulated results of ViX used in other vision transformer architectures such as LeViT, Compact Convolutional Transformer (CCT) and Convolutional vision Transformer (CvT)  for image classification with CIFAR-10 and Tiny ImageNet datasets is provided in the supplementary material.

We can see from Table~\ref{tab:cifLeV} that LeViX models using Performer (LeViP) and Nyströmformer (LeViN) outperform LeViT by about $1\%$ and $3\%$ respectively for classification of CIFAR-10 dataset. Even though all the models have a similar number of parameters, the LeViT model takes more than twice the memory and thrice the GPU RAM compared to LeViP and LeViN models. Similar performance gain is observed in the case of CCT and CvT, where Performer (CCP and CvP) and Nyströmformer (CCN and CvN) perform better than CCT and CvT using less than half its memory and one-third of GPU RAM. The same trend is observed in Table~\ref{tab:tinyLeV} for classification of Tiny ImageNet dataset where LeViP outperforms LeViT by $\approx 8\%$, CCN and CCP outperform CCT by $\approx 3\%$, and CvN and CvP outperform CvT by $\approx 12\%$. This shows that replacing quadratic attention with Performer and Nyströmformer attention will improve the performance in long sequence image classification using lesser memory and GPU RAM. These results indicate the possibility of training transformer models in vision with limited resources and less data using X-formers.

\begin{table*}[h]
\centering
%\begin{center}
%\resizebox{\textwidth}{!}{%
\begin{tabular}{lrrrrr}
\hline
\textbf{Model} & \textbf{Parameters} & \textbf{Size (MB)} & \textbf{Top-1 Accuracy (\%)} & \textbf{Top-5 Accuracy (\%)} & \textbf{GPU (GB)} \\ \hline
PiT & \textbf{897,034} & 129.74 & 63.49 & 96.95 & 14.7 \\
PiN & 897,562 & 65.81 & \textbf{69.41} & \textbf{97.68} & 5.7 \\
PiP & 898,570 & \textbf{64.09} & 65.83 & 97.28 & \textbf{5.2} \\ \hline
Hybrid PiT & \textbf{989,770} & 131.80 & 73.34 & \textbf{98.11} & 14.8 \\
Hybrid PiN &  990,298 & 67.87 & \textbf{74.00} & 98.03 & 5.7 \\
Hybrid PiP & 991,306 & \textbf{66.15} & 71.37 & 97.88 & \textbf{5.2} \\ \hline \\
\end{tabular}%
%}
%\end{center}
\caption{Comparison of performance by replacing attention with X-formers in PiT for classification of CIFAR-10 dataset}
\label{tab:PiX}
\end{table*}
% \usepackage{booktabs}
% \usepackage{graphicx}
\begin{table*}[h]
%\begin{center}
\centering

% \resizebox{\textwidth}{!}{%
\begin{tabular}{lrrr}
\hline
\textbf{Model}  & \textbf{Top-1 Accuracy (\%)} & \textbf{Top-5 Accuracy (\%)} & \textbf{GPU (GB)}\\ \hline
ViT & \textbf{69.32} & \textbf{97.68} & 14.4 \\
ViP &  60.74 & 95.62 & 5.2 \\
ViL & 55.88 & 94.61 & \textbf{2.7} \\
ViN & 68.36 & 97.28 & 5.2 \\ \hline
Hybrid ViT & \textbf{76.90} & \textbf{98.60} & 14.5 \\
Hybrid ViP & 76.86 & 98.37 & 5.2 \\
Hybrid ViL & 64.40 & 96.63 & \textbf{2.7} \\
Hybrid ViN & 74.38 & 97.64 & 5.2 \\ \hline \\
\end{tabular}%
% }
%\end{center}
\caption{Performance of ViX and Hybrid ViX models using Rotary Position Embedding on CIFAR-10 dataset}
\label{tab:RoPE}
\end{table*}

We also observe from Table~\ref{tab:PiX} that replacing the quadratic attention in pooling-based image transformer (PiT) by ViX improves its performance. Nyströmformer (PiN) and Performer (PiP) improves the accuracy by about $9 \%$ and $3 \%$ respectively using one-third the GPU RAM and half the memory compared to PiT. We also observe the similar trend from previous sections where a significant increase in accuracy was achieved by replacing the linear embedding layer with convolutional ones. An average increase of $10\%$ in classification accuracy was observed in Hybrid PiX models compared to PiX models.

%-------------------------------------------------------------------------

\subsection{ViX with Rotary Position Embedding}

Table \ref{tab:RoPE} shows that replacing the 1D learnable position embedding with Rotary Position Embedding (RoPE) further increases the performance of ViX and hybrid ViX models. We observed that using RoPE did not change the number of parameters in these models, but slightly increased the size and GPU usage. The trend of low GPU usage and memory consumption by ViX and hybrid ViX compared to ViT was observed again with RoPE.

%------------------------------------------------------------------------
\section{Conclusions and Discussion}

We showed that using linear self-attention mechanisms such as Performer and Nyströmformer in place of a vanilla attention mechanism with quadratic complexity in vision transformers can overcome a significant computational bottleneck that limits the application of transformers for long sequences formed by image pixels. More importantly, linear attention mechanisms in ViX can achieve image classification accuracy comparable to ViT while using far fewer computational resources, GPU RAM, and storage memory. This result was expected as linear attention mechanisms such as Nyströmformer and Performer were able to reach the performance of vanilla transformer using half the GPU in LRA benchmark tasks which included image classification~\cite{tay2020long,xiong2021nystromformer}. Our experiments confirm that replacing quadratic attention with these linear attention mechanisms in ViT architectures also leads to significant reduction in GPU RAM and computational costs without deteriorating their performance in image classification. This will enable the usage of transformer architectures for vision applications by practitioners who have resource constraints in terms of size of the GPUs (i.e., RAM and cores).

Additionally, we showed that replacing the initial fully connected embedding layer with the convolutional ones can independently improve the performance of vision transformers by providing an inductive bias that allows them to achieve higher accuracy with less training data. This result was obtained in parallel to ours by another group that showed similar trends by adding convolutional layers to vision transformer architectures, such as LeViT, CCT and CvT~\cite{graham2021levit,wu2021cvt,hassani2021escaping,heo2021rethinking}. We show that using convolutional layers to generate the embedding is enough to gain significant improvement in performance.

Furthermore, we showed that replacing the standard 1D learnable position embedding with rotary position embedding also showed independent improvement in classification accuracy. This trend was observed across different transformer architectures such as LeViT, CCT, CVT, and PiT when we implemented these changes in them. Alternative architectures that are efficient in token mixing, such as FNet and MLP mixer, also hold promise in vision applications as they require only one-tenth of the GPU RAM used by a vanilla ViT. Using ViX, FNet or MLP mixer will enable the use of deeper models with a fixed GPU RAM budget compared to vanilla ViT. Addition of convolutional layers for generating the embedding significantly improved the performance of FNet, but such an increase in accuracy was not observed with MLP mixer. This shows that making a hybrid MLP mixer does not justify the increase in space and parameter count. So, we do not recommend using convolutional layers in the MLP mixer.

Undoubtedly, further experiments with hyper-parameter tuning, using different positional embedding, alternative mixing models, and image datasets and tasks can lead to more insights for improving the efficiency and accuracy in low data, low GPU RAM regime. However, our results point to several significant directions for developing alternatives to purely convolutional architectures for vision. While the use of transformer architectures for vision has produced tantalizing results in terms of accuracy, these have come at orders of magnitude higher costs in terms of training data, floating-point operations, GPU RAM, hardware costs, and power consumption~\cite{li2020train}. This has put training transformer-type architectures out of the reach of all researchers, except those in a select few organizations with massive budgets, while their monetization is still in question. On the other hand, in spite of the theoretical guarantee of a fully connected neural network with a single hidden layer being able to model any function~\cite{kolmogorov1957representation}, proposals of neural architectures that exploit domain-specific inductive biases have resulted in usable increases in accuracy as well as decreases in computational and data requirements. Our work suggests that this quest is far from over, and architectural innovation can democratize the ability to train neural architectures from scratch with state-of-the-art performance.

\bibliographystyle{unsrtnat}
\bibliography{references}  %%% Uncomment this line and comment out the ``thebibliography'' section below to use the external .bib file (using bibtex) .

\end{document}